\documentclass[10pt,journal,compsoc]{IEEEtran}

\ifCLASSOPTIONcompsoc
  \usepackage[nocompress]{cite}
\else
  \usepackage{cite}
\fi

\ifCLASSINFOpdf
  \usepackage[pdftex]{graphicx}
\else
  \usepackage[dvips]{graphicx}
\fi

\usepackage{amsmath,amssymb,amsfonts}
\usepackage{amsthm}
\usepackage{mathrsfs}

\usepackage{array}
\usepackage{multirow}
\usepackage{booktabs}
\usepackage{subcaption}     % subfigures
\usepackage[export]{adjustbox}
\usepackage{stfloats}
\ifCLASSOPTIONcompsoc
  \usepackage[caption=false,font=footnotesize,labelfont=sf,textfont=sf]{subfig}
\else
  \usepackage[caption=false,font=footnotesize]{subfig}
\fi

\usepackage{url}
\usepackage{xcolor}
\begin{document}

\title{Frequency Selective Neural Networks as a Foundation Architecture for Time Series Learning}

% \author{First~Author,~\IEEEmembership{Member,~IEEE,}
%         Second~Author,~\IEEEmembership{Fellow,~IEEE,}
%         and~Third~Author,~\IEEEmembership{Member,~IEEE}% <-this % stops a space
\author{Hui~Huang,
        Ye~Sun,~\IEEEmembership{Member,~IEEE,}
        and~Shiyan~Hu,~\IEEEmembership{Senior Member,~IEEE}% <-this % stops a space
\IEEEcompsocitemizethanks{\IEEEcompsocthanksitem Hui Huang is with Coginfinite Technology, Germantown, MD, 20874.\protect\\
E-mail: huih.coginftech@gmail.com

\IEEEcompsocthanksitem Ye Sun is with the Department of Mechanical and Aerospace Engineering, Department of Electrical and Computer Engineering, University of Virginia\protect\\
E-mail: dzv7sg@virginia.edu

\IEEEcompsocthanksitem Shiyan Hu is with the Department of Data and Systems Engineering, University of Hong Kong\protect\\
E-mail:shiyanhu@hku.hk.}% <-this % stops an unwanted space
% \thanks{Manuscript received [Date]; revised [Date].}}}
}
\IEEEtitleabstractindextext{%
\begin{abstract}
Time-series data across physical and biological domains are fundamentally driven by complex, non-stationary oscillatory modes. While deep learning models, such as Convolutional Neural Networks (CNNs), Recurrent Neural Networks, and Transformers, have dominated sequential analysis, they remain fundamentally ``spectral-blind''. By mapping continuous physical waves into unconstrained spatial or discrete token spaces, these architectures suffer from severe spectral entanglement, acting as opaque black boxes that decouple predictive accuracy from physical reality. In this paper, we introduce the Frequency Selective Neural Network (FSNN), pioneering a foundation architecture guaranteeing physical interpretability without sacrificing expressive power of deep learning. FSNN addresses spectral entanglement by explicitly embedding the rigorous mathematics of advanced signal processing into its neural topology. Through a fully differentiable Wiener-like filter bank optimized via complex-domain backpropagation, FSNN autonomously discovers and isolates the precise physical modes of a given task. Extensive evaluations demonstrate that FSNN establishes state-of-the-art predictive performance, achieving 77.0\% average accuracy on the standard 10 multivariate UEA datasets and leading across all major metrics on the highly imbalanced PTB-XL clinical ECG benchmark. Crucially, in contrast to yielding abstract feature maps, FSNN converges directly on physically meaningful frequency bands, such as isolating the cardiac QRS complex, providing a highly scalable, interpretable paradigm for robust pattern recognition in complex temporal domains. Our code is available at: https://github.com/ad6174hhhh/FSNN.
\end{abstract}

}

\maketitle
\IEEEdisplaynontitleabstractindextext
\IEEEpeerreviewmaketitle

\section{Introduction}\label{sec:introduction}

\IEEEPARstart{T}{ime-series} analysis is a fundamental pillar of modern pattern recognition, critical to applications ranging from continuous healthcare monitoring to predictive maintenance and anomaly detection. The physical reality of these signals is that they are typically complex, non-stationary compositions of underlying oscillatory modes. In contrast to static periodic waves, the fundamental properties of these signals, such as their frequency, amplitude, and phase, evolve continuously over time. Distinct ``modes'' encapsulate the specific, recurring physical or biological patterns that govern these dynamic changes. However, the current standard in deep learning relies on architectures fundamentally disconnected from this physical reality \cite{Rudin2019-RUDSEB}.

Traditional deep learning architectures suffer from a fundamental representational mismatch. Convolutional Neural Networks (CNNs) \cite{lecun1989backpropagation} project time series into abstract spatial representations. Even strong temporal variants like InceptionTime \cite{fawaz2020inceptiontime} produce chaotic, overlapping frequency responses that fail to disentangle discrete physical phenomena---a problem known as spectral entanglement. Subsequent advancements in sequential modeling introduced Transformer-based architectures \cite{vaswani2017attention,wu2022flowformer}. Yet, these attention mechanisms treat sequential data primarily as discrete tokens, lacking a structural prior for continuous, oscillatory frequency modes. Concurrently, structural approaches such as TimesNet \cite{wu2023timesnet} attempt to leverage multi-periodicity by reshaping 1D signals into 2D tensors. Nevertheless, despite serving as a robust general-purpose backbone, TimesNet's 2D convolutional formulation inherently obscures the continuous frequency spectrum. Ultimately, the representational mismatch of mapping continuous physical waves into architectures designed for spatial pixels or discrete tokens results in models that act as ``spectral-blind'' black boxes, incapable of recovering physically interpretable components.

In contrast, classical digital signal processing has long respected the oscillatory nature of time series through the deployment of filter banks and optimal linear estimators. In diverse scientific and engineering disciplines, ranging from seismology and telecommunications to biomedical engineering, filter banks act as arrays of highly tuned band-pass filters designed to decompose complex signals into distinct, band-limited modes. This allows scientists to isolate specific physical phenomena, such as extracting alpha and beta waves from an electroencephalogram (EEG) or separating seismic waves from tectonic noise. When these physical signals are corrupted by interference, the Wiener filter \cite{wiener1949extrapolation} serves as the foundational mathematical framework for optimal linear filtering. By adaptively attenuating frequency bands where the noise-to-signal ratio is high, the Wiener formulation provides a mathematically rigorous, structurally smooth, and strictly band-limited representation of the true signal. The paramount advantage of these traditional signal processing techniques is their inherent physical interpretability that every algorithmic parameter, such as a filter's center frequency or bandwidth, corresponds directly to a tangible, real-world physical property.

However, these classical methods are fundamentally static, relying on fixed mathematical heuristics that cannot adapt dynamically to massive datasets. Although recent approaches have attempted to transition to learnable frequency parameters, they rely on overly rigid shapes that introduce mathematical artifacts (e.g., SincNet \cite{ravanelli2018speaker}, LEAF \cite{zeghidour2021leaf}), or completely unconstrained weights (e.g., FilterNet \cite{yi2024filternet}) that degrade into biologically implausible noisy representations.

The novelty of the Frequency Selective Neural Network (FSNN) lies in resolving this representational mismatch by explicitly prioritizing physical interpretability within an adaptable learning framework. Rather than abandoning structural priors or relying on chaotic parameters, FSNN natively enforces a fully differentiable, mathematically smooth Wiener-like prior. This physics-informed structural constraint prevents the network from learning chaotic, entangled noise. Instead, it ensures dynamic adaptability to massive datasets while forcing every learned parameter to isolate distinct, real-world physical frequencies, offering unprecedented scientific transparency. Ultimately, this transforms deep learning from a passive predictive black box into an active instrument for scientific discovery, empowering researchers to perform various scientific discoveries such as identifying novel physiological biomarkers, predicting uncharacterized mechanical faults, and formulating new physical hypotheses directly from the model’s converged weights.

The scientific significance of this study is that we are the first to propose a foundation architecture that inherently brings the exact, parameter-level interpretability of classical physics into the neural architecture by understanding time series as a composition of distinct physical modes, and retaining the dynamic optimization power of deep learning \cite{karniadakis2021physics}.
To this end, we introduce the Frequency Selective Neural Network. By rendering the strictly band-limited, smooth attenuation of Wiener-like filter banks end-to-end differentiable, FSNN transitions physical signal processing into a dynamic representation learning paradigm. This incremental structural constraint unifies the rigorous physical isolation of classical mathematics with the expressive capacity of modern artificial intelligence. Although the FSNN architecture is broadly applicable to diverse temporal tasks, we utilize time-series classification as our primary testbed to rigorously demonstrate its predictive efficacy and representational capabilities. Through rigorous evaluations on the UEA Time Series Classification Archive \cite{bagnall2018uea} and the PTB-XL electrocardiogram (ECG) dataset \cite{wagner2020ptb}, we show that FSNN consistently outperforms existing CNN, RNN, MLP, and Transformer baselines, establishing a new state-of-the-art in both predictive accuracy and model interpretability. Specifically, the core contributions of this paper are summarized as follows:
\begin{itemize}
    \item we propose FSNN as a new foundation architecture that represents time series as compositions of oscillatory modes; by introducing a dynamic, learnable filter bank, it unifies physical signal processing principles with end-to-end deep representation learning.
    \item We demonstrate that this physics-informed inductive bias directly translates to superior predictive accuracy. FSNN achieves a new state-of-the-art average accuracy of 77.0\% on standard multivariate UEA datasets and outperforms existing temporal and frequency-domain models on the highly imbalanced PTB-XL clinical ECG benchmark.
    \item Beyond reducing spectral entanglement, FSNN learns distinct frequency-selective filters with clear physical meaning. This provides explicitly interpretable representations that standard CNN, XceptionTime, and unconstrained FilterNet-style architectures do not offer, contributing a crucial step toward the physical interpretability of time-series learning.
\end{itemize}

\section{Related Work}\label{sec:related_work}

\subsection{Deep Learning for Time Series}
Deep representation learning for time series has historically relied on spatial and temporal priors adapted from computer vision and natural language processing. Over the last decade, the transition from classical statistical methods to deep neural networks has introduced robust mechanisms to model complex temporal dynamics. Foundational autoregressive models and recurrent architectures, such as LSTNet \cite{lai2018modeling}, were primarily employed to capture temporal dependencies by maintaining hidden state representations over time. However, as the complexity of real-world datasets has grown, classical recurrent methods have proven insufficient for modeling highly interconnected multivariate sequences. To address this, deep graph-evolution neural networks have been introduced to infer complex, multi-variate relationships \cite{jin2024survey}. These architectures assume that temporal data depends not only on intra-temporal relationships (past observations of a single variable) but also on inter-temporal dependencies across external variables, allowing models to significantly improve forecasting by tracking how multiple variables synchronously evolve \cite{spadon2021pay}.

Concurrently, convolutional architectures have been heavily adapted for sequence modeling. 1D-CNN variants like InceptionTime \cite{fawaz2020inceptiontime} apply multi-scale receptive fields to extract local features across varying temporal window sizes. However, these unconstrained spatial convolutions essentially act as arbitrary finite impulse response (FIR) filters \cite{stankovic2023convolutional}. Because their weights are optimized strictly through gradient descent without structural or spectral constraints, they frequently result in chaotic, overlapping frequency responses. This lack of spectral discipline leaves convolutional models highly prone to overfitting to high-frequency noise, making them particularly brittle when deployed on non-stationary, noisy signals found in real-world environments \cite{behroozi2025sensitivity}. 
To mitigate the above challenges and to improve the generalizability of temporal representations, the field has increasingly shifted toward self-supervised learning (SSL) paradigms. Modern SSL techniques for time series utilize generative, contrastive, and adversarial strategies to extract robust, invariant feature representations prior to downstream fine-tuning \cite{eldele2021self}. While these self-supervised frameworks represent a major methodological leap in tasks like forecasting, anomaly detection, and clustering, their underlying backbones still rely on standard temporal convolution or attention mechanisms, meaning they inherit the same spectral vulnerabilities.

% Standard deep representation learning for time series has historically relied on spatial and temporal priors adapted from computer vision and natural language processing. Models like LSTNet \cite{lai2018modeling} capture temporal dependencies, while 1D-CNN variants like InceptionTime \cite{fawaz2020inceptiontime} apply multi-scale receptive fields to extract local features. However, unconstrained spatial convolutions act as arbitrary finite impulse response (FIR) filters, frequently resulting in chaotic, overlapping frequency responses that are prone to overfitting to high-frequency noise \cite{stankovic2023convolutional}.
More recently, Transformer architectures have achieved significant success in sequence modeling. Variants such as Autoformer \cite{wu2021autoformer}, FEDformer \cite{zhou2022fedformer}, and Flowformer \cite{wu2022flowformer} are designed to capture long-range temporal dependencies. Because standard self-attention suffers from quadratic computational complexity with respect to sequence length, these models attempt to mitigate this bottleneck through various linearization, sparse attention matrices, and frequency-domain decomposition techniques. Furthermore, architectures like TimesNet \cite{wu2023timesnet} recently proposed transforming 1D series into 2D tensors based on multiple periods, allowing the direct application of highly optimized 2D vision backbones to capture inter-period and intra-period variations. However, all these models operate fundamentally in the spatial or temporal token domains, leaving them highly susceptible to high-frequency noise and spectral entanglement when processing continuous oscillatory signals.

\subsection{Classical Filter Banks and Wiener Filtering}
In the domain of digital signal processing (DSP), filter banks act as arrays of highly tuned band-pass filters designed to decompose complex, non-stationary signals into distinct, band-limited modes \cite{strang1996wavelets}. By mapping a one-dimensional time-domain signal into a two-dimensional time-frequency representation, filter banks effectively isolate specific oscillatory components, allowing for the targeted analysis of transient phenomena that would otherwise be obscured in raw temporal data. Classical systems rely heavily on fixed center frequencies and mathematically rigid bandwidths based on predefined heuristics \cite{huang2001spoken}. For example, wavelet transforms separate variations across different scales to linearize small deformations, providing sparse, orthogonal representations that are highly effective for signal compression and classical feature extraction \cite{mallat2016understanding}. Foundational work in this area, such as the wavelet packet decomposition introduced by Laine and Fan (1993) \cite{laine1993texture}, demonstrated that recursive subband filtering could effectively construct a tree-structured multiband extension to extract robust, invariant signatures from complex spatial or temporal data. 
Historically, these filter bank representations have been the standard front-end for deep learning architectures. For instance, computing filter bank features coupled with delta coefficients has been widely utilized to train robust Deep Neural Network (DNN) and Long Short-Term Memory (LSTM) models for automatic speech recognition systems \cite{hsu2017unsupervised}. More recently, research has actively explored utilizing mathematically disciplined filter banks, such as the Rational Dilated Wavelet Transform (RDWT) \cite{bayram2008overcomplete}, to act as a structured preprocessing layer that reliably suppresses localized noise before injecting signals into modern deep learning classifiers \cite{siino2025investigating}.

When these physical signals are corrupted by additive interference, channel distortion, or sensor noise, static filter banks are often insufficient for accurate reconstruction. In such scenarios, the Wiener filter provides an optimal, computationally efficient closed-form solution \cite{wiener1949extrapolation}. Operating structurally in the frequency domain, the Wiener filter is designed to minimize the mean square error (MSE) between the estimated underlying signal and the true corrupted signal \cite{hayes1996statistical}. As discussed by Yoo and Ahn \cite{yoo2014image}, achieving an optimal, linear least-mean-square estimate typically requires accurate prior knowledge of the power spectra of both the inherent noise and the original uncorrupted signal. By estimating these spectra, the algorithm inherently acts as an adaptive bandpass filter by balancing inverse filtering against noise amplification. This mathematically elegant, power-spectral-density-driven approach ensures that the structural integrity of the base signal is maintained while aggressively suppressing stochastic high-frequency noise. Despite their mathematical rigor, there is a fundamental structural disconnect between classical DSP algorithms and modern end-to-end deep learning pipelines. In their classical form, algorithms like the Wiener filter or deterministic wavelet filter banks remain largely static \cite{engel2020ddsp}. They are typically formulated as closed-form algebraic solutions or fixed mathematical transformations that do not natively support dynamic optimization via backpropagation based on task-specific loss functions.
\subsection{Learnable Spectral Frontends}
Recognizing the limitations of static preprocessing, recent research have increasingly sought to integrate digital signal processing principles directly into neural architectures. An important work in embedding physical inductive biases was the development of invariant scattering convolution networks by Bruna and Mallat \cite{bruna2013invariant}. Scattering networks compute a translation-invariant representation that is stable to local deformations by cascading fixed wavelet transform convolutions with nonlinear modulus and averaging operators. While scattering representations successfully incorporate higher-order moments for robust pattern discrimination, their reliance on predefined, rigid filter dictionaries prevents dynamic, gradient-based adaptation to specific datasets. Moreover, the necessary modulus operators inherently discard the phase information essential for precise signal reconstruction and cross-channel temporal alignment. Subsequent deep learning efforts have focused on rendering frequency-domain filters learnable.
For example, SincNet \cite{ravanelli2018speaker} parameterizes the first convolutional layer as a set of ideal band-pass filters (sinc functions), allowing the network to learn custom cut-off frequencies. LEAF \cite{zeghidour2021leaf} expanded this by utilizing parameterized Gabor filters. However, both methods rely on overly rigid mathematical shapes that introduce artifacts or are strictly confined to serving as 1D audio frontends. Conversely, recent frequency-domain architectures designed for general time series, such as FilterNet \cite{yi2024filternet}, utilize completely unconstrained complex weights in the spectral domain. Without a structural prior, these networks are prone to severe overfitting on noisy datasets and degrade into uninterpretable, biologically implausible representations.

\section{Proposed Methodology}\label{sec:methodology}

\subsection{Preliminaries}
\label{Pre}

\textbf{Filter Banks for Spectral Decomposition.}
In digital signal processing, a filter bank is an array of bandpass filters designed to decompose an input signal into multiple distinct frequency sub-bands, isolating specific spectral components for localized analysis. Traditionally, this paradigm has been central to various engineering domains, including audio compression, telecommunications, and biomedical signal analysis \cite{strang1996wavelets,tan2018digital}

The decomposition process, often referred to as the analysis bank, applies a set of $K$ filters with impulse responses $h_k(t)$ to an input signal $x(t)$. In the frequency domain, this is represented as:
\begin{equation}
X_k(\omega) = X(\omega)H_k(\omega), \quad k = 1, 2, \dots, K
\end{equation}
where $X(\omega)$ is the Fourier transform of the input signal, and $H_k(\omega)$ represents the frequency response of the $k$-th filter. In classical systems, such as the discrete wavelet transform or Mel-frequency filter banks used in speech recognition, the center frequencies $\omega_k$ and bandwidths of $H_k(\omega)$ are fixed based on predefined mathematical heuristics or psychoacoustic models \cite{huang2001spoken}. While highly effective for stationary signals with known characteristics, these static filter banks lack the adaptability required to dynamically isolate complex, non-stationary oscillatory modes without significant domain-specific manual tuning.

\textbf{Wiener Filtering.}
When decomposing a signal to extract meaningful modes from noise or interference, the Wiener filter serves as a foundational mathematical framework for optimal linear filtering \cite{wiener1949extrapolation}. Originally formulated to estimate a desired continuous-time stochastic process $s(t)$ from a corrupted observation $x(t) = s(t) + n(t)$ (where $n(t)$ is additive noise), the Wiener filter minimizes the mean square error (MSE) between the estimated signal and the true signal. Assuming the signal and noise are wide-sense stationary and uncorrelated, the optimal non-causal Wiener filter in the frequency domain is defined by the transfer function:

\begin{equation}
H_{wiener}(\omega) = \frac{S_{ss}(\omega)}{S_{ss}(\omega) + S_{nn}(\omega)}
\end{equation}

where $S_{ss}(\omega)$ and $S_{nn}(\omega)$ are the power spectral densities of the desired signal and the noise, respectively \cite{haykin2008adaptive}. A critical property of the Wiener formulation is its inherent structural constraint: it naturally acts as an adaptive bandpass filter that attenuates frequency bands where the noise-to-signal ratio is high, and preserves bands where the signal dominates. In the context of extracting a specific narrowband oscillatory mode centered at $\omega_k$ with a penalty constraint on its bandwidth $\alpha$, the Wiener-like spectral attenuation can be structurally idealized as:
\begin{equation}
H_k(\omega) = \frac{1}{1 + 2\alpha(\omega - \omega_k)^2}
\end{equation}

This structural prior ensures that the resulting filter is effectively band-limited and smooth, preventing spectral leakage and resisting wideband high-frequency noise.

Typically, both filter banks and Wiener filters have been deployed as static preprocessing steps or adaptive algorithms that rely on strict statistical assumptions. Recent advances in deep learning have attempted to integrate signal processing concepts directly into neural architectures \cite{ravanelli2018speaker,zeghidour2021leaf}. However, most current approaches either rely on unconstrained frequency-domain convolutions that easily overfit to noise, or utilize fixed algorithmic front-ends that cannot be optimized via backpropagation for a specific downstream task.
By rendering the classical mathematical formulations end-to-end differentiable, we successfully merge the optimal adaptability of deep learning with the strict, noise-resistant physical constraints of traditional signal processing in the following sections.

\subsection{System Architecture}

The core scientific innovation of the FSNN foundation architecture lies in transitioning classical signal processing, where every parameter possesses explicit physical meaning, into a dynamic, end-to-end differentiable neural layer. 

\begin{figure}[t]
    \centering
    \includegraphics[width=0.4\textwidth, height = 0.5\textwidth]{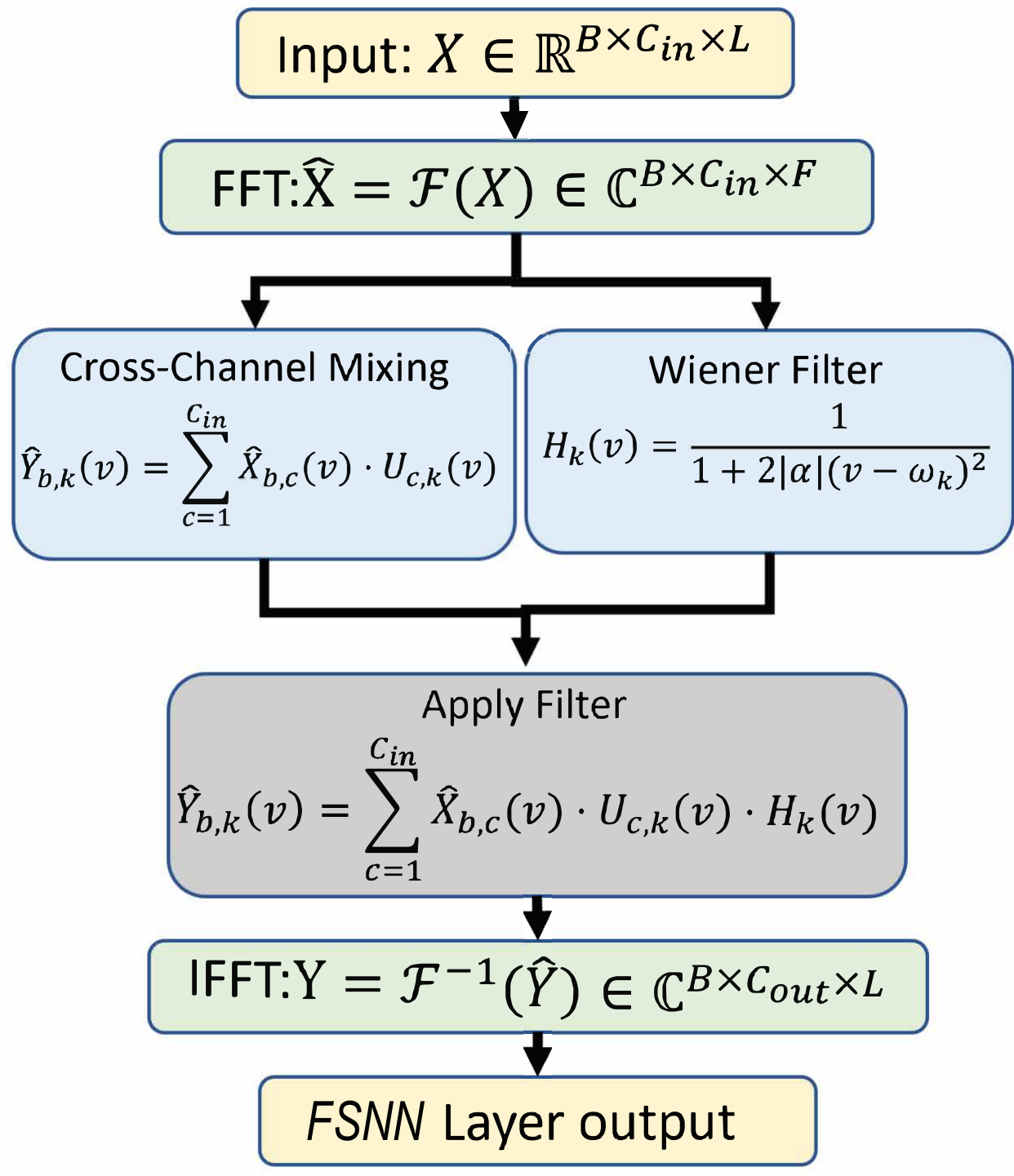}
    \caption{The core FSNN layer architecture. Input signals are transformed via RFFT, modulated by a differentiable Wiener-like filter bank and cross-channel mixing weights, and returned to the time domain via IFFT.}
    \label{fig:fsnn_layer}
\end{figure}

As illustrated in Figure \ref{fig:fsnn_layer}, unlike standard 1D convolutions that mix temporal signals indiscriminately in the time domain, the FSNN operates entirely in the spectral domain. It processes input signals by applying a set of mathematically constrained bandpass filters, parameterized by explicitly learnable center frequencies ($\omega$) and dynamic bandwidth penalties ($\alpha$). By enforcing this Wiener-like structural prior, the network is restricted from learning chaotic, biologically or physically implausible weights. Instead, it is forced to form clean, localized, bell-shaped spectral envelopes. 

During training, backpropagation serves as a guided physical constraint: when a particular frequency bin yields high discriminative value, the gradient adaptively steers the center frequency $\omega$ to that exact spectral location and tunes the bandwidth $\alpha$, either contracting to pinpoint a narrow rhythm or expanding to encapsulate a wider spectral trend. This mechanism shifts the deep learning process from abstract ``curve fitting'' to explicit ``mode discovery.'' 
\label{Methodology}
Building upon the principles of optimal linear filtering and filter banks discussed previously, we propose transitioning these classical static algorithms into a dynamic, task-driven representation. To this end, we introduce a novel differentiable layer within a neural network (FSNN). In this framework, the central frequencies $\omega_k$ and bandwidth parameters $\alpha$ are no longer static hyperparameters requiring manual tuning or complex heuristic search algorithms. Instead, they become trainable weights, initialized via physical heuristics (e.g., linear spacing) and refined via backpropagation to minimize the specific cross-entropy or regression loss of the target task. By making the spectral decomposition learnable, the network can ``tune'' its input filters to isolate the most discriminative spectral features, effectively performing task-driven signal processing. In cases of necessary interpretability, because the layer parameters correspond directly to physical quantities (center frequency and bandwidth), the learned model offers inherent transparency that one can inspect the converged $\omega_k$ values to understand exactly which physiological or mechanical rhythms the network is focusing on, bridging the gap between data-driven correlation and physical causation. 
% \begin{wrapfigure}{r}{0.5\textwidth}
%     \centering
%     \includegraphics[width=0.45\textwidth, height = 0.55\textwidth]{FSNN_layer.pdf}
%     \caption{FSNN layer.}
%     \label{fig:fsnn_layer}
% \end{wrapfigure}
In the following sections, we provide a rigorous theoretical derivation of the FSNN, position it within the broader context of differentiable signal processing, and present a comprehensive empirical evaluation demonstrating its superiority over both traditional machine learning pipelines and their interpretabilities.

\subsection{Theoretical Formulation and Forward Pass}
As shown in Figure \ref{fig:fsnn_layer}, given the input to the FSNN layer be a time-series tensor $X\in\mathbb{R}^{B\times C_{in}\times L}$, where $B$ is the batch size, $C_{in}$ is the number of input channels, and $L$ is the sequence length. The forward pass maps this input to an output tensor $Y\in\mathbb{R}^{B\times C_{out}\times L}$, where $C_{out}$ represents the number of extracted modes (or output channels).

\textbf{Frequency Domain Transformation.}
The input tensor is projected into the frequency domain using the Real Fast Fourier Transform (RFFT) to leverage the Hermitian symmetry of real-valued signals, halving the computational cost:
\begin{equation}
    \hat{X} = \mathcal{F}(X) \in \mathbb{C}^{B\times C_{in}\times F}
\end{equation}
where $F = \lfloor L/2 \rfloor + 1$ is the number of frequency bins, and $\mathcal{F}(\cdot)$ denotes the RFFT operator. Let $\nu \in [0, 0.5]$ represent the normalized frequency grid associated with these bins. With the frequency domain transformation, the computational efficiency remains strictly bounded and scalable, as detailed in the theoretical complexity analysis in Appendix. 

\textbf{Differentiable Filter Bank Construction.} To impose a physics-informed inductive bias, we explicitly construct a parameterized Wiener-like filter bank. We define key sets of parameters to enable dynamic frequency scaling and focus:
\begin{itemize}
\item \textbf{Learnable Center Frequencies ($\boldsymbol{\omega}$)}: The center frequencies of the Wiener-like attenuation term, $\boldsymbol{\omega} \in \mathbb{R}^{C_{out}}$, are defined as learnable parameters. They are initialized by being spaced evenly across the frequency spectrum. During training, backpropagation adjusts these frequencies to focus the filters on the most discriminative spectral bands for the given dataset. This enables the network to isolate the specific oscillatory rhythms that dictate class separation.
\item \textbf{Learnable Bandwidth Penalty ($\alpha_k$)}: Rather than treating the bandwidth constraint as a static hyperparameter, it is defined as a learnable parameter $\alpha \in \mathbb{R}^{C_{out}}$ providing a dedicated bandwidth penalty $\alpha_k$ for each mode $k$. In the forward pass, its absolute value $|\alpha_k|$ is utilized to guarantee the penalty term remains strictly positive. This flexibility allows the network to automatically discover and ``zoom in'' on the frequency ranges that contain the most class-relevant information, adapting to wideband signals by decreasing the penalty or isolating highly localized frequencies by increasing it.
\item \textbf{Learnable Spectral Weight Tensor ($U\in\mathbb{C}^{C_{in}\times C_{out}\times F}$)}: The primary learnable parameters are the complex-valued weight tensor. Each weight tensor   is a vector of the same length as the signal's spectrum and is learned directly. These weight tensors define the ideal frequency-domain shape that the layer seeks to extract from the input signal for each mode. They are initialized with small random values and optimized via backpropagation.
\end{itemize}
For each output mode $k \in \{1, \dots, C_{out}\}$, the spectral transfer function $H_k(\nu)$ is computed as:
\begin{equation}
    H_{k}(\nu)=\frac{1}{1+2|\alpha_{k}|(\nu-\omega_{k})^{2}}
\end{equation}
This formulation guarantees that the learned filters are effectively band-limited, structurally smooth, and inherently resistant to wideband noise.

\textbf{Spectral Modulation and Cross-Channel Mixing.} Unlike standard 1D convolutions that mix channels in the time domain, FSNN mixes channels directly in the frequency domain using the learnable complex-valued weight tensor $U$. The modulation for the $b$-th batch and $k$-th output mode at frequency bin $\nu$ is formalized as:
\begin{equation}
    \hat{Y}_{b,k}(\nu) = \sum_{c=1}^{C_{in}} \hat{X}_{b,c}(\nu) \cdot H_k(\nu) \cdot U_{c,k}(\nu)
\end{equation}
Here, $H_k(\nu)$ acts as the structural prior isolating the specific frequency band, while the complex weights $U_{c,k}(\nu)$ allow the network to apply necessary amplitude scaling and phase shifts to optimally combine the input channels.

Finally, the modulated spectra are transformed back into the time domain via the Inverse RFFT:
\begin{equation}
    Y = \mathcal{F}^{-1}(\hat{Y}) \in \mathbb{R}^{B\times C_{out}\times L}
\end{equation}
This output $Y$ can subsequently be passed through standard non-linearities (e.g., ReLU, GELU) and stacked to form deep, hierarchical FSNN architectures.

\subsection{Backpropagation and Gradient Flow}
The end-to-end learning capability of FSNN relies on the differentiability of the Wiener filter formulation. During training, a task-specific objective function $\mathcal{L}$ (e.g., cross-entropy for classification) is minimized. The gradients flow backward through the IFFT operator to the modulated spectrum $\hat{Y}$. The update mechanism for the complex mixing weights $U_{c,k}(\nu)$ follows standard automatic differentiation:
\begin{equation}
    \frac{\partial\mathcal{L}}{\partial U_{c,k}(\nu)}=\sum_{b=1}^{B}\left(\frac{\partial\mathcal{L}}{\partial\hat{Y}_{b,k}(\nu)}\right)^*\cdot\hat{X}_{b,c}(\nu)\cdot H_{k}(\nu)
\end{equation}
Crucially, the gradients driving the evolution of both the center frequencies $\omega_k$ and the bandwidth penalty $\alpha$ expose the task-driven nature of the FSNN architecture:

\textbf{Center Frequency Gradient}: By applying the chain rule, the gradient with respect to $\omega_k$ is:
\begin{equation} 
\begin{split} 
\frac{\partial\mathcal{L}}{\partial\omega_{k}} & =\Re\left\{\sum_{b=1}^{B}\sum_{c=1}^{C_{\text{in}}}\sum_{\nu}\left(\frac{\partial\mathcal{L}}{\partial\hat{Y}_{b,k}(\nu)}\right)^*\cdot\hat{X}_{b,c}(\nu)\cdot U_{c,k}(\nu) \right. \\ 
& \quad \left. \cdot\frac{\partial H_{k}(\nu)}{\partial\omega_{k}}\right\} 
\end{split} 
\end{equation}
Taking the partial derivative of the Wiener transfer function $H_k(\nu)$ with respect to $\omega_k$ yields:
\begin{equation}
    \frac{\partial H_k(\nu)}{\partial \omega_k} = \frac{4|\alpha_k|(\nu - \omega_k)}{\left(1 + 2|\alpha_k|(\nu - \omega_k)^2\right)^2}
\end{equation}    
This derivative acts as a directional force. If a specific frequency bin $\nu$ is highly discriminative for the classification task (resulting in a large upstream gradient $\frac{\partial \mathcal{L}}{\partial \hat{Y}}$), the term $4|\alpha|(\nu - \omega_k)$ proportionally pulls the center frequency $\omega_k$ towards $\nu$. Thus, the gradient flow dynamically shifts the filter banks to converge exactly on the spectral modes that separate the classes.

\textbf{Bandwidth Penalty Gradient}:
Similarly, the learnable bandwidth penalty $\alpha$ is updated to adapt the filter's narrowness. The gradient with respect to $\alpha$ is calculated via:
\begin{equation}
\begin{split} 
\frac{\partial\mathcal{L}}{\partial\alpha_{k}} & =\Re\left\{\sum_{b=1}^{B}\sum_{c=1}^{C_{in}}\sum_{\nu}\left(\frac{\partial\mathcal{L}}{\partial\hat{Y}_{b,k}(\nu)}\right)^*\cdot\hat{X}_{b,c}(\nu)\cdot U_{c,k}(\nu) \right. \\
& \quad \left. \cdot\frac{\partial H_{k}(\nu)}{\partial\alpha_{k}}\right\}
\end{split} 
\end{equation}
where:
\begin{equation}
    \frac{\partial H_k(\nu)}{\partial \alpha_k} = -\text{sgn}(\alpha_k) \frac{2(\nu - \omega_k)^2}{\left(1 + 2|\alpha_k|(\nu - \omega_k)^2\right)^2}
\end{equation}
This gradient dictates the dynamic frequency scaling. It allows the network to adaptively expand or compress the passband of the filters. If discriminative information is distributed across a wider spectral band, the gradient updates will decrease $|\alpha|$, whereas highly localized frequency features will drive $|\alpha|$ higher to form a narrow, highly selective filter. This resolves the parameter selection bottleneck natively within gradient descent.

\begin{figure*}[t]
    \centering
    \begin{subfigure}[t]{0.32\textwidth}
        \centering
        \includegraphics[width=\linewidth]{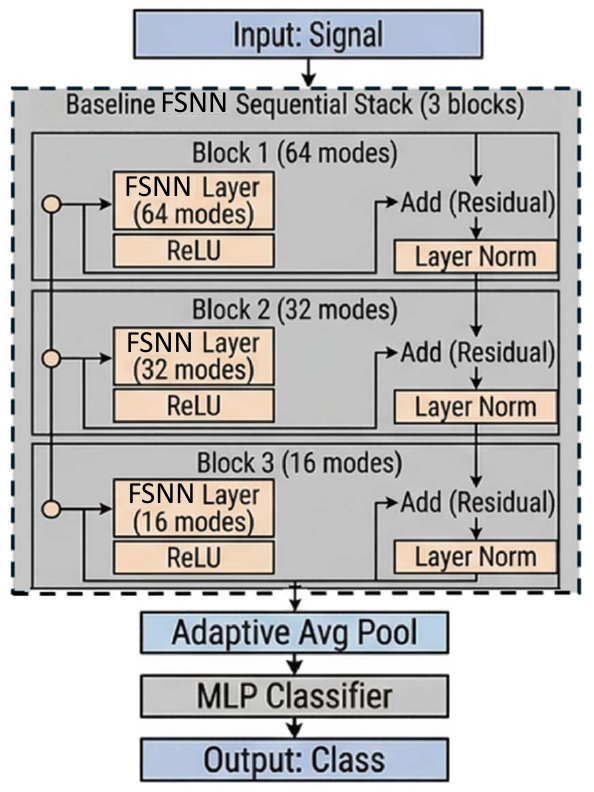}
        \phantomsubcaption
        \label{fig:sys1}
        \caption*{(a)}
    \end{subfigure}
    \hfill
    \begin{subfigure}[t]{0.32\textwidth}
        \centering
        \includegraphics[width=\linewidth,height=1.35\linewidth]{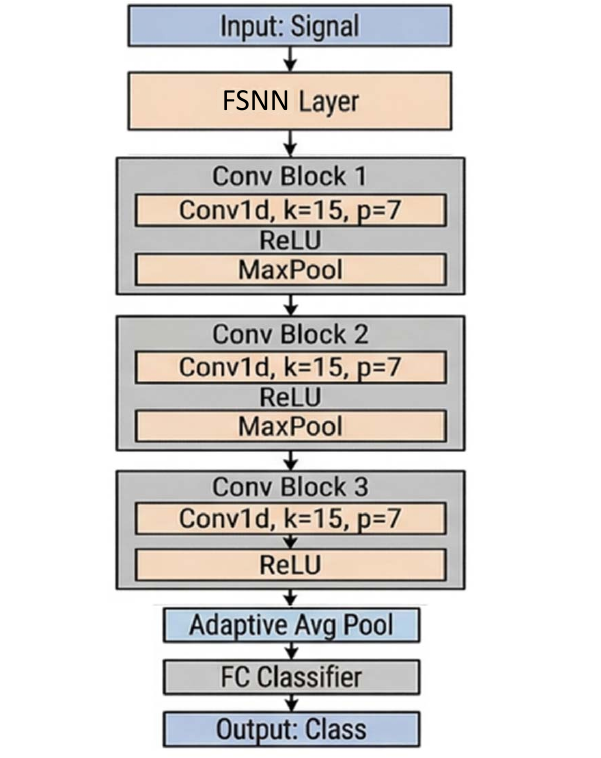}
        \phantomsubcaption
        \label{fig:sys2}
        \caption*{(b)}
    \end{subfigure}
    \hfill
    \begin{subfigure}[t]{0.32\textwidth}
        \centering
        \includegraphics[width=\linewidth,height=1.35\linewidth]{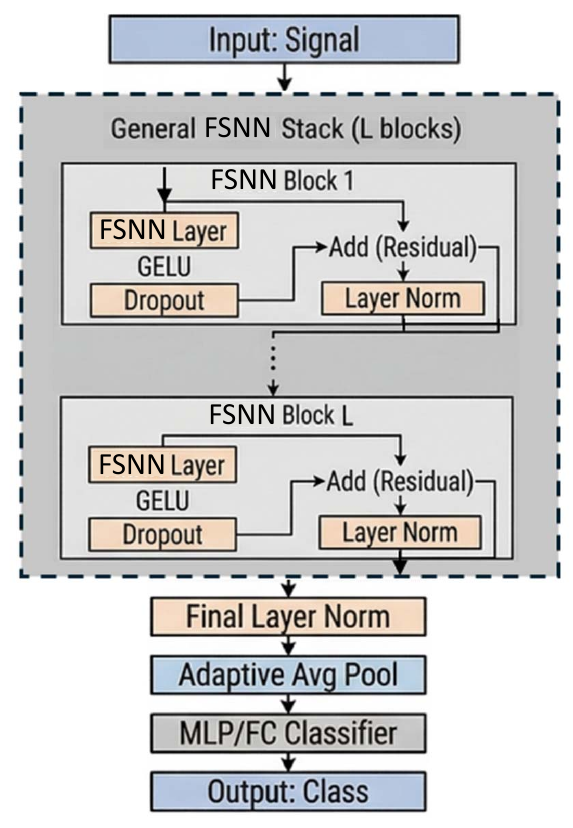}
        \phantomsubcaption
        \label{fig:sys3}
        \caption*{(c)}
    \end{subfigure}
    \caption{Three FSNN integration strategies evaluated in this study. (a) A standalone hierarchical feature extractor. (b) An interleaved spectral front-end for localized temporal convolutions. (c) A scalable, task-agnostic backbone embedded within isotropic residual blocks.}
    \label{fig:FSNN-systems}
\end{figure*}

\subsection{Complexity Analysis}
To demonstrate the efficiency of the FSNN architecture, particularly in comparison to the $\mathcal{O}(L^2 \cdot d)$ time complexity of canonical self-attention, we formalize the asymptotic computational and memory complexity of a single FSNN layer for an input sequence of length $L$ with $C_{in}$ input channels and $C_{out}$ output modes.

\textbf{Time Complexity}: The forward pass consists of three primary operations. First, the Real Fast Fourier Transform (RFFT) projects the input into the frequency domain, requiring $\mathcal{O}(C_{in} \cdot L \log L)$ operations. Second, the cross-channel mixing and element-wise filter application in the frequency domain (over $F \approx L/2$ bins) takes $\mathcal{O}(C_{in} \cdot C_{out} \cdot L)$ time. Finally, the Inverse RFFT maps the modulated spectra back to the time domain, taking $\mathcal{O}(C_{out} \cdot L \log L)$ time. Therefore, the overall time complexity of the FSNN layer is:$$\mathcal{O}((C_{in} + C_{out}) \cdot L \log L + C_{in} \cdot C_{out} \cdot L)$$Because $C_{in}$ and $C_{out}$ are typically small constants representing channel or mode dimensions, the sequence-length time complexity is strictly $\mathcal{O}(L \log L)$. This quasi-linear scaling is significantly more efficient than quadratic self-attention, making FSNN highly suitable for exceptionally long time-series sequences.

\textbf{Space Complexity}: The memory footprint is primarily dictated by the learnable complex weight tensor $U \in \mathbb{C}^{C_{in} \times C_{out} \times F}$ and the intermediate frequency-domain activations. Storing the parameters requires $\mathcal{O}(C_{in} \cdot C_{out} \cdot L)$ space. The intermediate activations for a batch size $B$ require $\mathcal{O}(B \cdot \max(C_{in}, C_{out}) \cdot L)$ space. Thus, the overall space complexity is linearly proportional to the sequence length, denoted as $\mathcal{O}(L)$.

\subsection{Integration Strategies with Deep Learning Architectures} 
The inherent flexibility of the FSNN layer allows it to serve as a core computational unit within diverse deep learning topologies. To evaluate its generalizability, we conceptualize three distinct architectural integrations as examples. Figure \ref{fig:sys1} shows a basic FSNN architecture similar to a CNN model. This architecture treats the FSNN layer as an autonomous, hierarchical feature extractor, completely replacing temporal convolutions. The model stacks multiple FSNN layers sequentially, separated by non-linear activation functions (e.g., ReLU). The configuration expands the channel dimensions progressively (e.g., mapping $1 \rightarrow 64 \rightarrow 32 \rightarrow 16$ modes), representing a deep decomposition process where broad frequency modes are subsequently refined into highly granular sub-modes. The extracted representations are then globally pooled and projected via a multi-layer perceptron (MLP) for classification.

% \begin{figure*}[t]
%     \centering
%     \subfloat[Baseline FSNN]{\includegraphics[width=0.32\linewidth]{sys1.pdf}%
%     \label{fig:sys1}}
%     \hfil
%     \subfloat[Hybrid FSNN-CNN]{\includegraphics[width=0.32\linewidth]{sys2.pdf}%
%     \label{fig:sys2}}
%     \hfil
%     \subfloat[Residual FSNN]{\includegraphics[width=0.32\linewidth]{sys3.pdf}%
%     \label{fig:sys3}}
%     \caption{Three FSNN integration strategies evaluated in this study. (a) A standalone hierarchical feature extractor. (b) An interleaved spectral front-end for localized temporal convolutions. (c) A scalable, task-agnostic backbone embedded within isotropic residual blocks.}
%     \label{fig:FSNN-systems}
% \end{figure*}

Figure \ref{fig:sys2} shows a hybird integration of FSNN and CNN. This strategy capitalizes on the complementary strengths of frequency-domain filtering and time-domain convolutions. The architecture interleaves FSNN layer with standard 1D Convolutional Neural Network (CNN) layers ($\text{FSNN} \rightarrow \text{Conv1d} \rightarrow \text{RELU}$). Here, the FSNN layer acts as a global, physics-informed spectral filter, disentangling overlapping oscillatory components into clean intrinsic modes. The subsequent Conv1d layers apply localized temporal receptive fields to extract precise morphological features from these isolated modes. Mathematically, the block transformation leverages $\mathcal{F}_{CNN}(\sigma(\mathcal{F}_{FSNN}(X)))$, providing a highly discriminative, noise-resilient representation pipeline.

% Figure \ref{fig:sys2} shows a hybrid integration of FSNN and CNN. This strategy capitalizes on the complementary strengths of frequency-domain filtering and time-domain convolutions. The architecture interleaves FSNN layer with standard 1D Convolutional Neural Network (CNN) layers, utilizing the FSNN to extract global, physics-informed spectral modes while subsequent Conv1D layers capture precise localized morphological features $\mathcal{F}_{CNN}(\sigma(\mathcal{F}_{FSNN}(X)))$.

Figure \ref{fig:sys3} displays a more complicated FSNN model. Designed as a scalable, task-agnostic backbone for broad time-series analysis , this architecture embeds the FSNN layer into an isotropic residual block, mirroring the structural topology of modern Transformers. An initial projection layer maps the raw input to a constant hidden dimension $d_{model}$. The core FSNN Block processes the signal by applying the FSNN layer, followed by a GELU activation, Dropout regularization, a skip connection, and Layer Normalization.
By leveraging memory-efficient techniques for cross-channel mixing within the FSNN layer , this architecture safely scales to deep stacks. Except for classification tasks, one can design and add more downstream tasks through specialized output heads, utilizing sequence-level global average pooling for classification, or specific temporal and channel projections for imputation and forecasting.

% Figure \ref{fig:sys3} displays a more complicated FSNN model. Designed as a scalable, task-agnostic backbone for broad time-series analysis, this architecture embeds FSNN within an isotropic residual block (featuring GELU, Dropout, and Layer Normalization) to form a scalable, complex and task-agnostic model. Mirroring modern Transformer topologies, this memory-efficient design safely supports deep stacking and diverse downstream applications, such as forecasting and imputation.

\begin{table*}[b]
\centering
\caption{Accuracy results (\%) for the time-series classification with UEA datasets. Abbreviations are used for the transformers. The standard diviation is within 0.1\%. Best results are bolded.}
\label{tab:tsc_comp}
\resizebox{\textwidth}{!}{%
\begin{tabular}{l*{18}{c}}
\toprule
Datasets / Models & \multicolumn{2}{c}{Classical} & \multicolumn{2}{c}{RNN} & \multicolumn{2}{c}{TCN} & \multicolumn{9}{c}{Transformers} & \multicolumn{2}{c}{MLP} & \textbf{FSNN} \\
\cmidrule(lr){2-3}\cmidrule(lr){4-5}\cmidrule(lr){6-7}\cmidrule(lr){8-16}\cmidrule(lr){17-18}\cmidrule(lr){19-19}
& DTW & Rocket & LSTNet & LSSL & TCN & TimesNet & Trans. & Re. & In. & Pyra. & Auto. & Station. & FED. & ETS. & Flow. & DLinear & LightTS & \textbf{(Ours)} \\
\midrule
EthanolConcentration & 32.3 & \textbf{45.2} & 39.9 & 31.1 & 28.9 & 35.7 & 32.7 & 31.9 & 31.6 & 30.8 & 31.6 & 32.7 & 31.2 & 28.1 & 33.8 & 32.6 & 29.7 & 36.1 \\
FaceDetection & 52.9 & 64.7 & 65.7 & 66.7 & 52.8 & \textbf{68.6} & 67.3 & \textbf{68.6} & 67.0 & 65.7 & 68.4 & 68.0 & 66.0 & 66.3 & 67.6 & 68.0 & 67.5 & 67.3 \\
Handwriting & 28.6 & 58.8 & 25.8 & 24.6 & 53.3 & 32.1 & 32.0 & 27.4 & 32.8 & 29.4 & 36.7 & 31.6 & 28.0 & 32.5 & 33.8 & 27.0 & 26.1 & \textbf{60.1} \\
Heartbeat & 71.7 & 75.6 & 77.1 & 72.7 & 75.6 & 78.0 & 76.1 & 77.1 & \textbf{80.5} & 75.6 & 74.6 & 73.7 & 73.7 & 71.2 & 77.6 & 75.1 & 75.1 & 80.0 \\
JapaneseVowels & 94.9 & 96.2 & 98.1 & 98.4 & 98.9 & 98.4 & 98.7 & 97.8 & 98.9 & 98.4 & 96.2 & 99.2 & 98.4 & 95.9 & 98.9 & 96.2 & 96.2 & \textbf{99.2} \\
PEMS-SF & 71.1 & 75.1 & 86.7 & 86.1 & 68.8 & \textbf{89.6} & 82.1 & 82.7 & 81.5 & 83.2 & 82.7 & 87.3 & 80.9 & 86.0 & 83.8 & 75.1 & 88.4 & 86.1 \\
SelfRegulationSCP1 & 77.7 & 90.8 & 84.0 & 90.8 & 84.6 & 91.8 & 92.2 & 90.4 & 90.1 & 88.1 & 84.0 & 89.4 & 88.7 & 89.6 & \textbf{92.5} & 87.3 & 89.8 & 89.1 \\
SelfRegulationSCP2 & 53.9 & 53.3 & 52.8 & 52.2 & 55.6 & 57.2 & 53.9 & 56.7 & 53.3 & 53.3 & 50.6 & 57.2 & 54.4 & 55.0 & 56.1 & 50.5 & 51.1 & \textbf{61.7} \\
SpokenArabicDigits & 96.3 & 71.2 & \textbf{100.0} & \textbf{100.0} & 95.6 & 99.0 & 98.4 & 97.0 & \textbf{100.0} & 99.6 & \textbf{100.0} & \textbf{100.0} & \textbf{100.0} & \textbf{100.0} & 98.8 & 81.4 & \textbf{100.0} & \textbf{100.0} \\
UWaveGestureLibrary & 90.3 & \textbf{94.4} & 87.8 & 85.9 & 88.4 & 85.3 & 85.6 & 85.6 & 85.6 & 83.4 & 85.9 & 87.5 & 85.3 & 85.0 & 86.6 & 82.1 & 80.3 & 90.6 \\
\midrule
Average Accuracy & 67.0 & 72.5 & 71.8 & 70.9 & 70.3 & 73.6 & 71.9 & 71.5 & 72.1 & 70.8 & 71.1 & 72.7 & 70.7 & 71.0 & 73.0 & 67.5 & 70.4 & \textbf{77.0} \\
\bottomrule
\end{tabular}%
}
\end{table*}

\section{Experiments}\label{sec:experiments}

\subsection{Representational Superiority on Benchmark Datasets}
To evaluate the representational capacity of the FSNN, we conducted an extensive evaluation across 10 diverse multivariate datasets from the UEA Time Series Classification Archive \cite{bagnall2018uea}. We compared the full Residual FSNN backbone against a comprehensive suite of sequence modeling paradigms, including The benchmark models include: (1) RNN-based models, LSTNet \cite{lai2018modeling} and LSSL \cite{gu2022lssl}; (2) TCN-based models, TCN \cite{bai2018empirical} and TimesNet \cite{wu2023timesnet}; (3) MLP-based models, DLinear \cite{zeng2023transformers} and LightTS \cite{zhang2022less}; (4) Transformer-based models, Reformer \cite{kitaev2020reformer}, Informer \cite{zhou2021informer}, Pyraformer \cite{liu2021pyraformer}, Autoformer \cite{wu2021autoformer}, Non-stationary Transformer \cite{liu2022nonstationary}, FEDformer \cite{zhou2022fedformer}, ETSformer \cite{woo2022etsformer}, and Flowformer \cite{wu2022flowformer}; and other types of models such as DTW \cite{berndt1994using} and Rocket \cite{dempster2020rocket}.  
To ensure fair comparison with prior work  \cite{wu2023timesnet,wu2022flowformer}, we evaluate on the same 10 multivariate datasets from the UEA Time Series Classification Archive \cite{bagnall2018uea} that are commonly used in related studies such as TimesNet and Flowformer. These datasets span diverse application domains, including medical diagnosis, gesture recognition, action recognition, and audio classification. We follow the same preprocessing protocol as in prior work  \cite{zerveas2021transformer}. Our model is implemented in PyTorch, and all experiments are conducted on a single NVIDIA RTX 4060 Ti (8GB) GPU. we partitioned 20\% of the standard training data to serve as an independent validation set. We train the models using the Adam optimizer with an initial learning rate of $10^{-3}$, a batch size of 32, 10 random seeds of weight initializations for individual runs, and early stopping with a patience of 20 epochs. We use grid search to identify suitable hyperparameters, including the number of encoder layers and filter settings.

\begin{figure*}[t]
    \centering
    \includegraphics[width=0.9\textwidth,height= 0.35\textwidth]{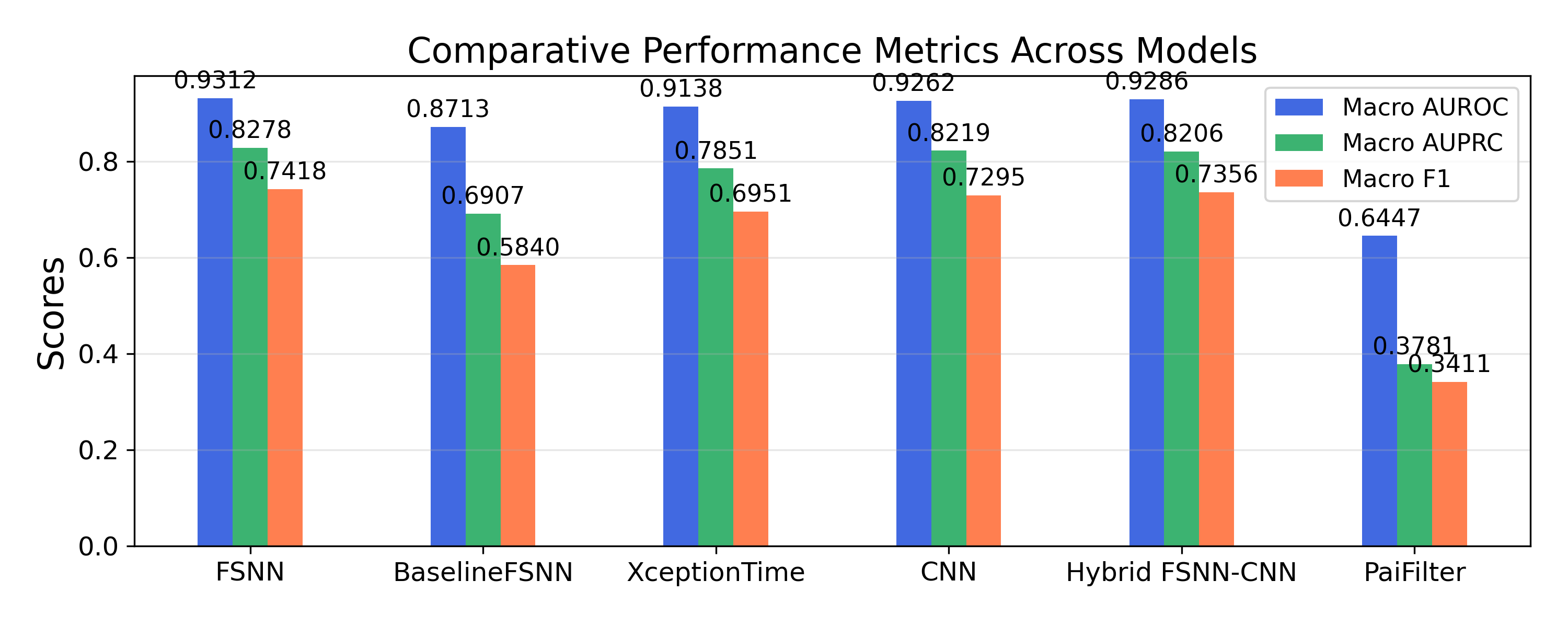}
    \caption{Performance comparison on PTB-XL dataset across baseline and proposed architectures.}
    \label{fig:performance-comparison}
\end{figure*}

The quantitative classification results across the 10 UEA datasets are summarized in our performance comparison in Table \ref{tab:tsc_comp}. The empirical evidence clearly demonstrates that the proposed FSNN (Fig. \ref{fig:sys3}) achieves state-of-the-art performance, consistently outperforming the vast majority of baseline architectures across diverse domains.

When compared to the extensive lineup of Transformer-based models (such as FEDformer, ETSformer, and Flowformer), FSNN showcases superior discriminative capability. While Transformers often struggle with high-frequency noise and require complex attention-linearization strategies to handle long sequences, the FSNN inherently sidesteps these issues. By leveraging its differentiable Wiener filter bank, FSNN directly isolates the underlying oscillatory modes that define the class boundaries, providing a stronger and more noise-resilient inductive bias than self-attention.

Furthermore, FSNN demonstrates a clear advantage over recent structural innovations like TimesNet. While TimesNet relies on transforming 1D series into 2D tensors to capture multi-periodicity via 2D-CNN backbones, FSNN achieves superior accuracy operating entirely within the 1D spectral domain. By learning the optimal center frequencies and bandwidth constraints directly from the data, FSNN captures complex intra- and inter-period variations without the need for manual 2D folding or the computational overhead of large convolutional kernels. Ultimately, the results validate that parameterizing traditional signal processing into an end-to-end deep learning layer establishes a highly effective, general-purpose foundation for time series classification.

Table \ref{tab:tsc_comp} reports the comparative results against strong benchmark methods. FSNN achieves the best overall performance, with an average accuracy of 77.0\%, surpassing previous state-of-the-art baselines, including TimesNet (73.6\%), Flowformer (73.0\%), and Rocket (72.5\%). These results demonstrate the excellent time-series modeling capability of FSNN.
This superior performance validates the scientific premise of the architecture: providing a deep neural network with an inductive bias tailored to the physical nature of oscillatory signals vastly improves its representational power. While Transformer models treat time series as arbitrary sequences and must rely on vast parameter counts and complex attention maps to infer global structure, FSNN directly isolates the underlying oscillatory modes that define class boundaries. By aligning the network's architecture with the fundamental physics of the data, FSNN achieves superior predictive performance without the need for unconstrained, biologically implausible feature spaces.

\subsection{Physical Interpretability and Clinical ECG Classification}
While the previous section establishes FSNN's superiority against broad sequence models (Transformers, RNNs), our evaluation on the PTB-XL dataset specifically focuses on comparing frequency-domain and convolutional feature extractors. This focused benchmark allows us to directly contrast the physical meaning and noise resilience of the learned filters (e.g., FSNN vs. PaiFilter and standard CNNs) in a highly imbalanced, noisy clinical setting. PTB-XL dataset is a large-scale clinical database of 12-lead ECG signals and is a highly imbalanced multi-label classification across five diagnostic super-classes: Normal (NORM), Myocardial Infarction (MI), ST/T Change (STTC), Conduction Disturbance (CD), and Hypertrophy (HYP). 
We compared three variants of our proposed architecture (BaselineFSNN (Fig. \ref{fig:sys1}), Hybrid FSNN-CNN (Fig. \ref{fig:sys2}), and FSNN (Fig. \ref{fig:sys3})) against established temporal models (1D-CNN, XceptionTime) and a recent frequency-domain model (PaiFilter/FilterNet).

As shown in Figure~\ref{fig:performance-comparison}, the quantitative results of the leading performance with Macro AUROC (0.9312), Macro AUPRC (0.8278) and Macro F1-score (0.7418) by FSNN indicate that the integration of physical signal decomposition into the network produces a robust feature space while maintaining strong predictive performance. In addition, our Hybrid FSNN-CNN achieves an AUROC of 0.9286 and a Macro F1-score of 0.8206, matching the baseline 1D-CNN in overall effectiveness and slightly surpassing it in AUROC (0.9262) and F1 (0.7295). This near-parity suggests that using the FSNN layer as a spectral front-end preserves the discriminative information needed by the downstream convolutional backbone. In addition, BaselineFSNN, which relies solely on hierarchical mode reduction without convolutional mixing, still achieves a competitive AUROC of 0.8713. These results show that mathematically constrained signal decomposition can serve as a strong standalone backbone for complex classification tasks.
The advantage of the FSNN framework is even more pronounced when compared with existing frequency-domain architectures. Our models substantially outperform PaiFilter, which struggles on this challenging multi-label clinical task, achieving only 0.6447 AUROC and 0.3411 F1. This large gap underscores an important architectural point: the Wiener-filter constraints built into the FSNN layer provide a strong inductive bias that stabilizes learning under high inter-patient variability and noisy ECG recordings. In contrast, unconstrained frequency models such as PaiFilter lack this physical structure, which can lead to severe performance degradation in real-world biomedical settings.

A key limitation of conventional deep learning models in healthcare is their black-box behavior. Although CNNs and Transformers can achieve high predictive accuracy, they provide limited insight into the physiological basis of their decisions. FSNN addresses this limitation through intrinsic, physically grounded interpretability. Visualization of learned frequency responses reveals a clear contrast between FSNN and competing architectures in how signal information is represented and utilized.

\begin{figure*}[htbp]
    \centering
    \setlength{\lineskip}{0pt}
    \begin{subfigure}[t]{0.8\textwidth}
        \centering
        (a) \hspace{5pt}
        \includegraphics[width=0.7\linewidth,height=0.25\linewidth, valign=c]{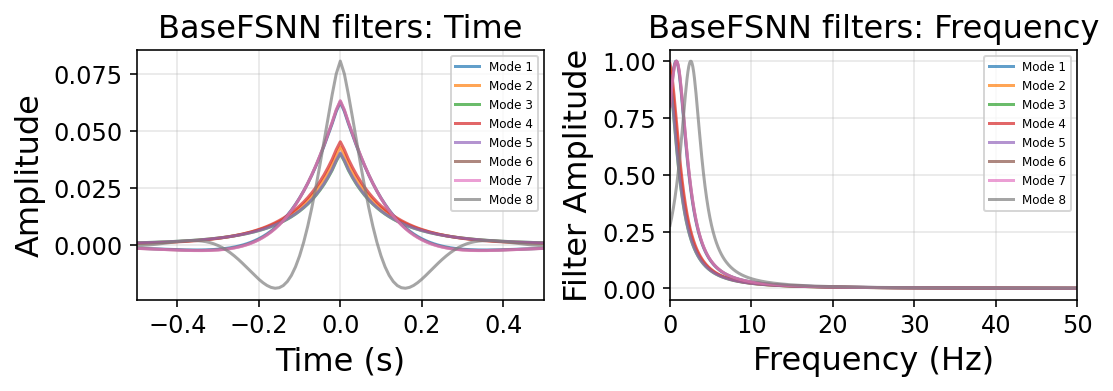}
        % \vspace{-0.5em}
        \phantomsubcaption
        \label{fig:baseFSNN_filters}
        % \caption*{(a)}
    \end{subfigure}\hfill
    \begin{subfigure}[t]{0.8\textwidth}
        \centering
        (b) \hspace{5pt}
        \includegraphics[width=0.7\linewidth,height=0.25\linewidth, valign=c]{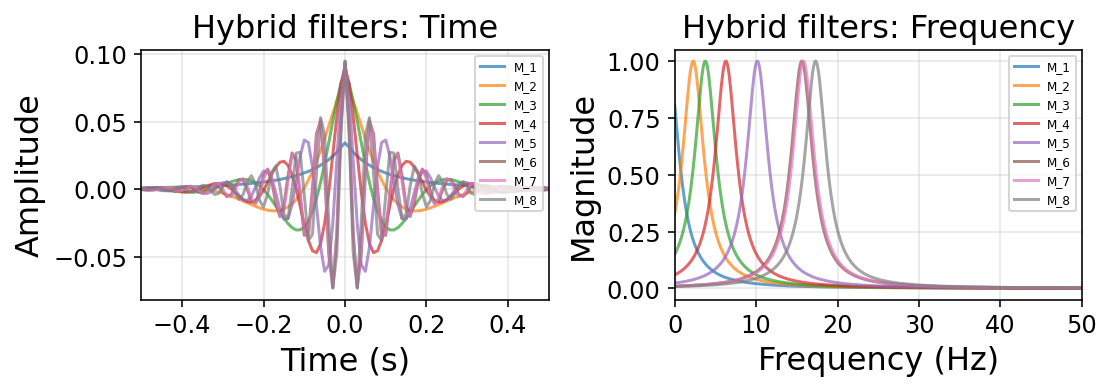}
        % \vspace{-0.5em}
        \phantomsubcaption
        \label{fig:hybrid_filters}
        % \caption*{(b)}
    \end{subfigure}\\
    \begin{subfigure}[t]{0.8\textwidth}
        \centering
        (c) \hspace{5pt}
        \includegraphics[width=0.7\linewidth,height=0.25\linewidth, valign=c]{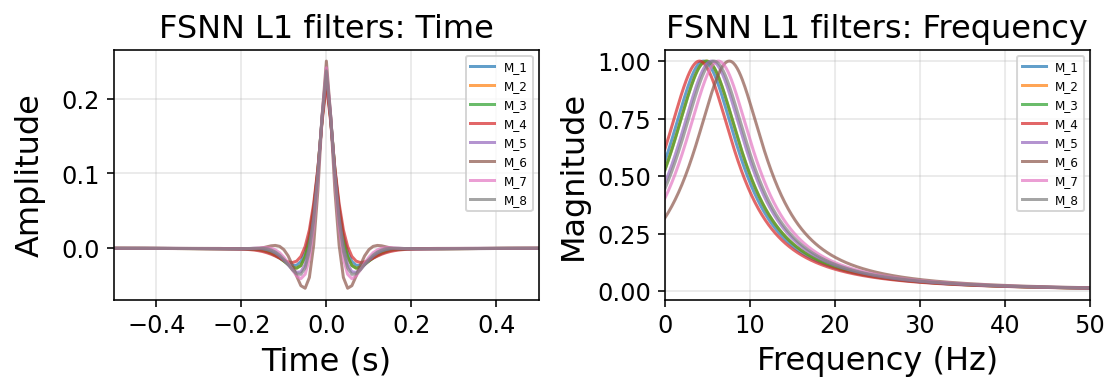}
        % \vspace{-0.5em}
        \phantomsubcaption
        \label{fig:fsnn_filters}
        % \caption*{(c)}
    \end{subfigure}\hfill
    \begin{subfigure}[t]{0.8\textwidth}
        \centering
        (d) \hspace{5pt}
        \includegraphics[width=0.7\linewidth,height=0.25\linewidth, valign=c]{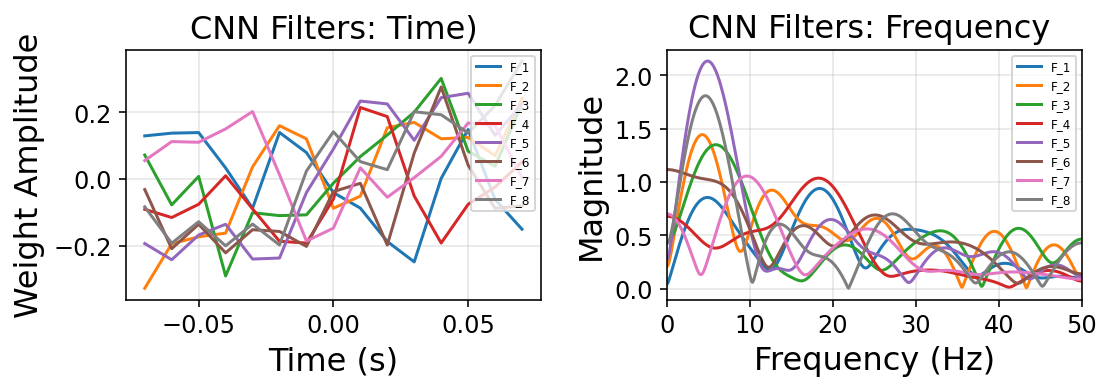}
        % \vspace{-0.5em}
        \phantomsubcaption
        \label{fig:cnn_filters}
        % \caption*{(d)}
    \end{subfigure}\\
    \begin{subfigure}[t]{0.8\textwidth}
        \centering
        (e) \hspace{5pt}
        \includegraphics[width=0.7\linewidth,height=0.25\linewidth, valign=c]{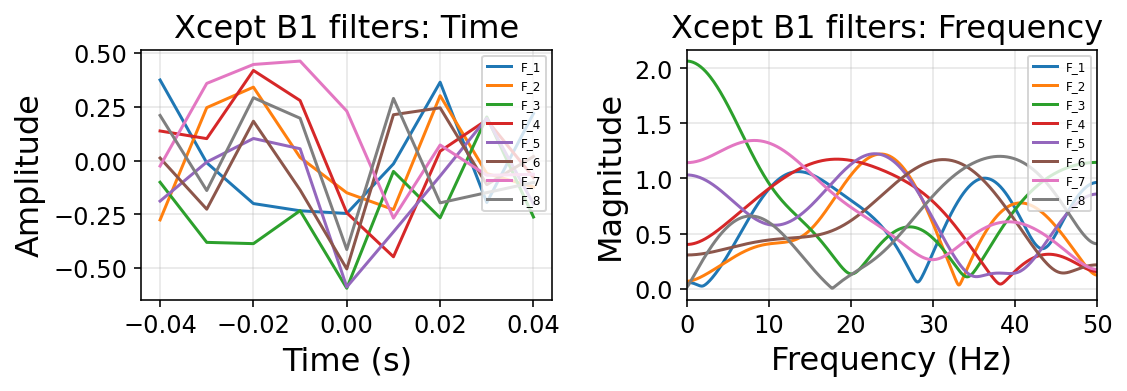}
        % \vspace{-0.5em}
        \phantomsubcaption
        \label{fig:xcept_filters}
        % \caption*{(e)}
    \end{subfigure}\hfill
    \begin{subfigure}[t]{0.8\textwidth}
        \centering
        (f) \hspace{5pt}
        \includegraphics[width=0.7\linewidth,height=0.25\linewidth, valign=c]{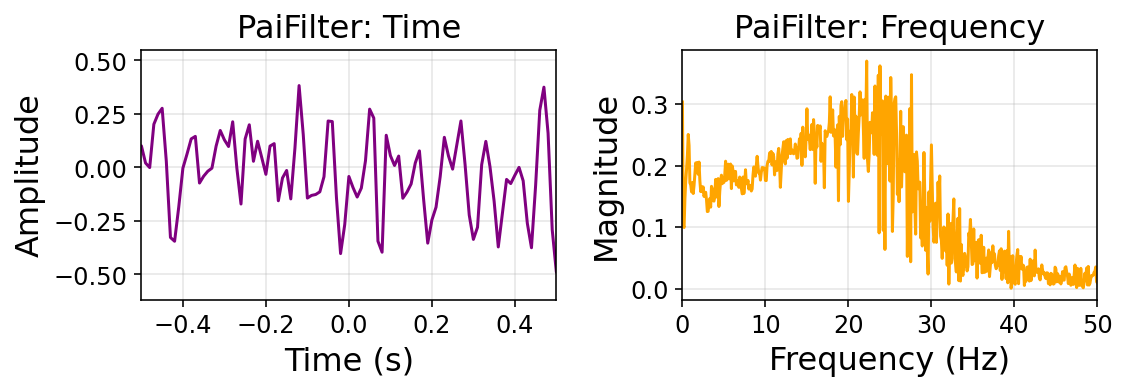}
        % \vspace{-0.5em}
        \phantomsubcaption
        \label{fig:paifilter_filters}
        % \caption*{(f)}
    \end{subfigure}
    \vspace{-0.8em}
    \caption{Learned filters with time and frequency domain: (a) baseline FSNN layer 1, (b) hybrid FSNN-CNN layer 1, (c) FSNN layer 1, (d) CNN layer 1, (e) XceptionTime, and (f) PaiFilter/FilterNet.}
    \label{fig:filters}
\end{figure*}

As shown in Figure \ref{fig:filters}, the learned parameters of FSNN form clean, localized, bell-shaped spectral envelopes. Instead of fitting noisy or diffuse spectra, the center frequencies converge to physiologically meaningful cardiac bands. In particular, displayed in Figure \ref{fig:baseFSNN_filters}, \ref{fig:hybrid_filters}, and \ref{fig:fsnn_filters}, FSNN and its variant isolate low-frequency components ($<5$ Hz) associated with P-wave and T-wave morphology, which are relevant for STTC detection. Moreover, separate modes are allocated to the 10--25 Hz range in the first layer of hybrid FSNN-CNN model, capturing the high-energy QRS complex that is important for myocardial infarction detection.
By contrast, the frequency responses of standard 1D-CNN and XceptionTime kernels are often jagged and multi-band as shown in Figure \ref{fig:cnn_filters} and \ref{fig:xcept_filters}. They show substantial spectral leakage across the higher frequency spectrum plotted (e.g., 20–40 Hz). In real-world ECG signals, this extended high-frequency range is increasingly dominated by electromyographic artifacts rather than stable cardiac structure. This pattern suggests that conventional CNN-based models may partially rely on stochastic, patient-specific noise cues instead of robust physiological markers.
Finally, in Figure \ref{fig:paifilter_filters}, although PaiFilter visualizations show broad full-spectrum coverage, they lack the strict mathematical inductive bias imposed by FSNN's Wiener formulation. As a result, PaiFilter does not cleanly separate intrinsic modes and tends to produce overlapping, noisy spectral bands that mix artifacts with true cardiac components, which is consistent with its weaker classification performance. Overall, by learning continuous mode-adaptive frequency filters under explicit spectral constraints, FSNN suppresses high-frequency clinical noise and aligns its representations more closely with physiological knowledge.

\subsection{Discussion}
As highlighted in our introduction, standard deep learning architectures suffer from spectral blindness and spectral entanglement, while classical signal processing techniques remain rigid and unadaptable. We overcome these structural challenges through three core contributions.
First, FSNN establishes a new foundation architecture that represents time series as compositions of oscillatory modes rather than discrete spatial or temporal tokens. We achieved this by introducing a dynamic, learnable filter bank that embeds a mathematically rigorous Wiener-like filter directly into the neural topology. This unifies the strict physical constraints of classical signal processing with the expressive, end-to-end optimization of deep representation learning, successfully bridging the historical disconnect between static algorithmic preprocessing and unconstrained machine learning.
Second, we demonstrated that this physics-informed inductive bias translates directly into state-of-the-art predictive performance. By enforcing a smooth, band-limited structural prior in the frequency domain, FSNN naturally resists high-frequency noise overfitting and effectively disentangles overlapping spectral phenomena. This structural mechanism enabled the architecture to reach 77.0\% average accuracy across 10 diverse UEA datasets, surpassing robust baselines like TimesNet and Flowformer, and to lead all major evaluation metrics on the highly imbalanced PTB-XL ECG benchmark.
Finally, FSNN directly addresses the ``black-box'' opacity that plagues traditional CNNs, Transformers, and unconstrained FilterNet-style architectures. We achieve physical interpretability by ensuring that the network's optimized parameters, center frequencies and bandwidths, correspond strictly to tangible, real-world physical properties. Rather than relying on abstract spatial feature maps, FSNN learns distinct, frequency-selective filters with clear physical meaning. As demonstrated in our clinical ECG evaluation, this allows researchers to directly inspect the learned weights to recover underlying physiological causation, representing a crucial step toward inherently interpretable time-series learning.

While FSNN establishes a strong physics-informed foundation for time series, it possesses limitations that outline clear directions for future research. The mathematically rigorous Wiener filter prior assumes the presence of distinct, stable oscillatory modes within the signal. For phenomena characterized by highly chaotic, impulsive, or strictly non-periodic signals (e.g., sudden mechanical impacts or purely random walk financial data), this strictly band-limited prior may prove too restrictive compared to unconstrained temporal convolutions. Furthermore, because the learnable complex weight tensor is defined over the discrete frequency spectrum, its dimensionality is currently tied to the sequence length. Future work will focus on integrating dynamic, complex-domain spectral interpolation to natively handle highly variable forecasting horizons, and extending the FSNN layer to incorporate multivariate cross-channel attention for complex spatial-temporal graphs.

\section{Conclusion}\label{sec:conclusion}
In this work, we introduced the FSNN, a new foundation architecture that brings the rigor of advanced signal processing into the flexible domain of deep representation learning. By transitioning classical, static filter banks into an end-to-end optimization of deep learning, FSNN overcomes the "spectral blindness" of traditional CNNs and Transformers. Our experiments demonstrate that FSNN achieves state-of-the-art accuracy on general time-series benchmarks and highly imbalanced clinical tasks like the PTB-XL dataset. 
More importantly, this approach represents a necessary philosophical shift toward physics-informed machine learning. Rather than relying on unconstrained parameters to map complex time series into abstract, uninterpretable latent spaces, FSNN forces
the network to learn through continuous, mode-adaptive Wiener-like filters. It learns explicitly in terms of center frequencies, bandwidths, and oscillatory modes, naturally isolating physically meaningful frequency bands, such as the QRS complex in ECGs, while inherently rejecting high-frequency noise. FSNN establishes a powerful, transparent, and robust "white-box" alternative for the potential interpretability of time-series analysis in high-stakes domains.

% We introduced the Frequency Selective Neural Network (FSNN), successfully unifying mathematically rigorous signal decomposition with end-to-end deep learning. By transitioning optimal linear filtering into a fully differentiable neural layer, FSNN overcomes the fundamental ``spectral blindness'' of traditional CNNs and Transformers. The model inherently learns explicitly in terms of center frequencies, bandwidths, and oscillatory modes. This physics-informed approach yields superior predictive accuracy on generalized benchmarks (77.0\% on UEA) and provides unprecedented physical interpretability on clinical data. FSNN establishes that enforcing a strict mathematical prior does not limit deep learning's expressive capacity, but rather unlocks robust representation learning. Future work will focus on integrating dynamic, complex-domain spectral interpolation to handle chaotic, non-periodic signals natively.

\bibliographystyle{IEEEtran}
\bibliography{references}

\end{document}